\documentclass[11pt]{article}

\usepackage[preprint]{acl}

\usepackage{times}
\usepackage{latexsym}

\usepackage[T1]{fontenc}
\usepackage[utf8]{inputenc}

\usepackage{microtype}

\usepackage{inconsolata}

\usepackage{booktabs}
\usepackage{graphicx}
\usepackage{xcolor}
\usepackage{enumitem}
\usepackage{multirow}
\usepackage{alltt}
\usepackage[textsize=scriptsize]{todonotes}
\usepackage{amsfonts}
\usepackage{tcolorbox}
\tcbuselibrary{most}
\definecolor{burntorange}{HTML}{BF5700}

\newtcolorbox{imagebox}[2][]{%
    enhanced,
    breakable,
    nobeforeafter,
    title={#2},
    coltitle=white,
    colframe=burntorange!40!white,
    colback=burntorange!3!white,
    colbacktitle=burntorange!90!black,
    fonttitle=\bfseries,
    sharp corners,
    boxrule=0.5pt,
    boxsep=1pt,
    left=6pt,
    right=6pt,
    top=0pt,
    bottom=0pt,
    #1
}

\newtcolorbox{modelsection}[2][]{%
    enhanced,
    breakable,
    title={#2},
    nobeforeafter,
    coltitle=black,
    colframe=burntorange!40!white,
    colback=burntorange!3!white,
    colbacktitle=burntorange!20!white,
    fonttitle=\bfseries,
    sharp corners,
    boxrule=0.5pt,
    boxsep=2pt,
    left=6pt,
    right=6pt,
    #1
}
\usepackage{lineno}
\usepackage{amsmath}
\definecolor{darkblue}{rgb}{0, 0, 0.5}
\hypersetup{colorlinks=true, citecolor=darkblue, linkcolor=darkblue, urlcolor=darkblue}

\title{Visually Grounded Self-Reflection for Vision-Language Models \\via Reinforcement Learning}

\author{Liyan Tang$^{\textbf{*}\spadesuit}, $Fangcong Yin$^{\textbf{*}\diamondsuit}$,
\textbf{Greg Durrett}$^{\diamondsuit}$ \\
$^{\spadesuit}$The University of Texas at Austin $^{\diamondsuit}$New York University  \\
\url{lytang@utexas.edu}, \url{{fy666,gdurrett}@nyu.edu}
}

\begin{document}

\maketitle

\begin{abstract}

Large vision-language models can reason over multimodal inputs by generating textual chains of thought (CoT). A key capability exhibited in CoT reasoning is self-reflection: revisiting earlier decisions and correcting previous errors. However, existing LVLMs often fail to properly attend to \emph{visual} inputs during reflection, limiting their ability to translate feedback into grounded corrections, especially for out-of-distribution images. To address this issue, we propose a novel reinforcement learning training framework \textbf{VRRL}, with two components explicitly designed to elicit visually grounded self-reflection. First, we randomly mask trajectory prefixes during training to emphasize recovery from incorrect intermediate predictions rather than making early mistakes. Second, we introduce buffered roll-ins from an experience replay buffer to expose the model to diverse failure states that it must learn to correct. We evaluate our approach on visual grounding tasks involving tables and charts, as well as spatial navigation benchmarks. While off-the-shelf and conventionally fine-tuned models degrade substantially under distribution shift, our method substantially improves average out-of-distribution accuracy over standard RL and reflection-oriented fine-tuning baselines by using self-reflection effectively.\footnote{$^{\textbf{*}}$Equal contribution. Our code and data are available at: \url{https://github.com/fc2869/VRRL}}
\end{abstract}

\section{Introduction}

Large vision-language models (LVLMs) can solve complex multimodal reasoning tasks over text and images by generating CoT traces \citep{zhang2024multimodal,hu2024visual}. Textual reasoning models use CoT to improve performance through a number of cognitive skills \cite{gandhi2025cognitive}, among them \emph{self-reflection}. Self-reflection is the act of a model reasoning about the correctness of a candidate answer or solution step, then potentially correcting it or revising it if an error was made \citep{madaan2023selfrefine,Guo2025}. However, this skill remains under-developed in LVLMs. 
Prior work suggests that a key bottleneck stems from the modality gap \citep{yi2024bridge}:
LVLMs often fail to attend to relevant visual tokens \citep{ma2026understanding} and struggle to translate visual evidence into grounded corrective behaviors \citep{huang2025highresolutionvisualreasoningmultiturn, zhang2026mirror}, particularly for complex or out-of-distribution (OOD) images that differ substantially from pre-training distributions. Achieving visual self-reflection in LVLMs requires additional post-training.

\begin{figure*}
    \centering
    \includegraphics[width=\linewidth]{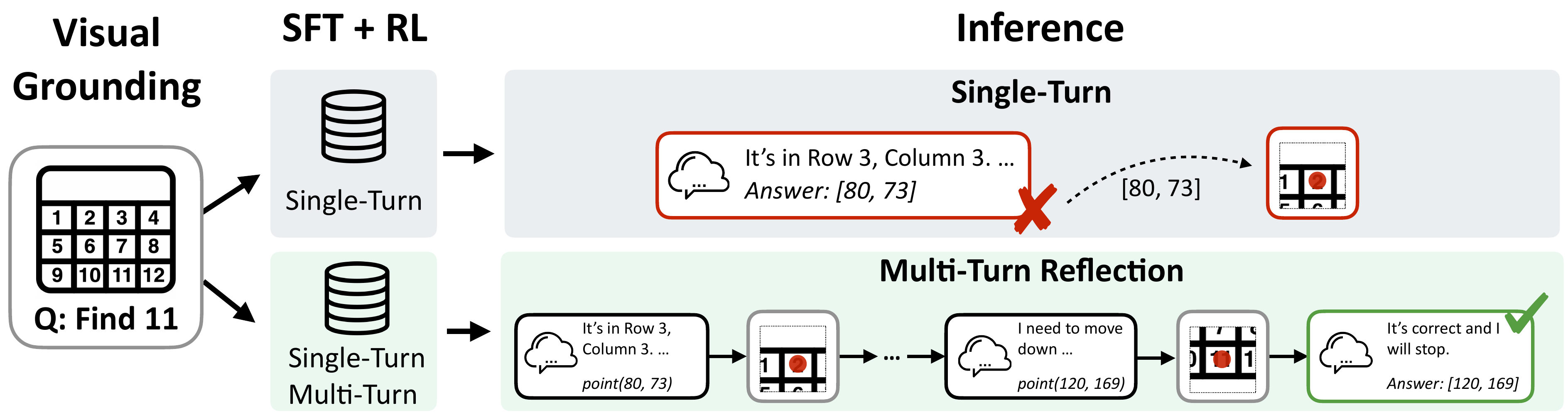}
    \caption{A motivating example of multi-turn reflection for visual grounding. In this illustrative task, the goal is to find the pixel coordinate for the value ``11'' in a table image. In a single turn, an LVLM can reason over text in the instruction but may fail to provide accurate pixel-level grounding output. However, if given its prediction as visual feedback (a red dot on the image), the LVLM can reflect to iteratively correct its inaccurate grounding and determine when to stop and give an answer.}
    \label{fig:multi_vs_single}
\end{figure*}

To address this limitation, we first use supervised fine-tuning (SFT) to teach the model the basic structure of multi-turn visual feedback and then propose a novel \textbf{V}isual \textbf{R}eflection \textbf{RL} (VRRL) training recipe explicitly designed to instill reflective error-correction from such feedback. During RL, VRRL combines two complementary strategies that expose the model to diverse error-recovery scenarios. First, for newly generated on-policy trajectories, we introduce \textbf{Random Turn Masking}, which computes policy updates only on randomly selected suffixes of a rollout. This teaches the model to learn to correct errors while masking the steps which may have led to those errors.
Second, we introduce a \textbf{Buffered Roll-In} strategy that samples historical ``mistake prefixes'' from a replay buffer of past failures and asks the model to continue the rollout and correct previous errors. By explicitly training the model to reason over multi-turn visual feedback, these strategies improve both the robustness and OOD generalization of visual reasoning.

We evaluate VRRL on two tasks where a VLM acts in an environment that provides visual feedback: (1) \emph{Visual grounding}, where the model predicts the coordinates of a queried visual element in an image; and (2) \emph{Spatial navigation}, where the model solves maze-like navigation tasks from visual inputs. For both tasks, we first establish basic self-reflection capabilities through SFT, and then enhance them using several RL baselines and our proposed method under the same training distribution \citep{Guo2025,kimiteam2025kimik15scalingreinforcement, zhai2024finetuning, feizi2026grounding}. We evaluate whether each approach generalizes its reflective correction behavior to OOD settings.

Our experiments show that off-the-shelf LVLMs struggle to generalize OOD under direct prompting, and that prompt engineering alone fails to elicit meaningful self-reflection, often leading to repetitive behaviors that degrade performance. Explicit training for self-reflection improves OOD generalization, while VRRL further outperforms standard RL and existing reflection-oriented fine-tuning methods by inducing visually grounded error correction more effectively.

The main contributions of this work are as follows: (1) We demonstrate that visually grounded self-reflection is an effective mechanism for improving LVLM robustness under distribution shifts, outperforming non-reflective and weakly grounded reflection baselines. (2) We introduce VRRL, a novel RL framework that combines Random Turn Masking and Buffered Roll-In to train models to recover from diverse intermediate errors, leading to stronger OOD generalization across visual feedback environments.

\begin{figure*}
    \centering
    \includegraphics[width=0.7\linewidth]{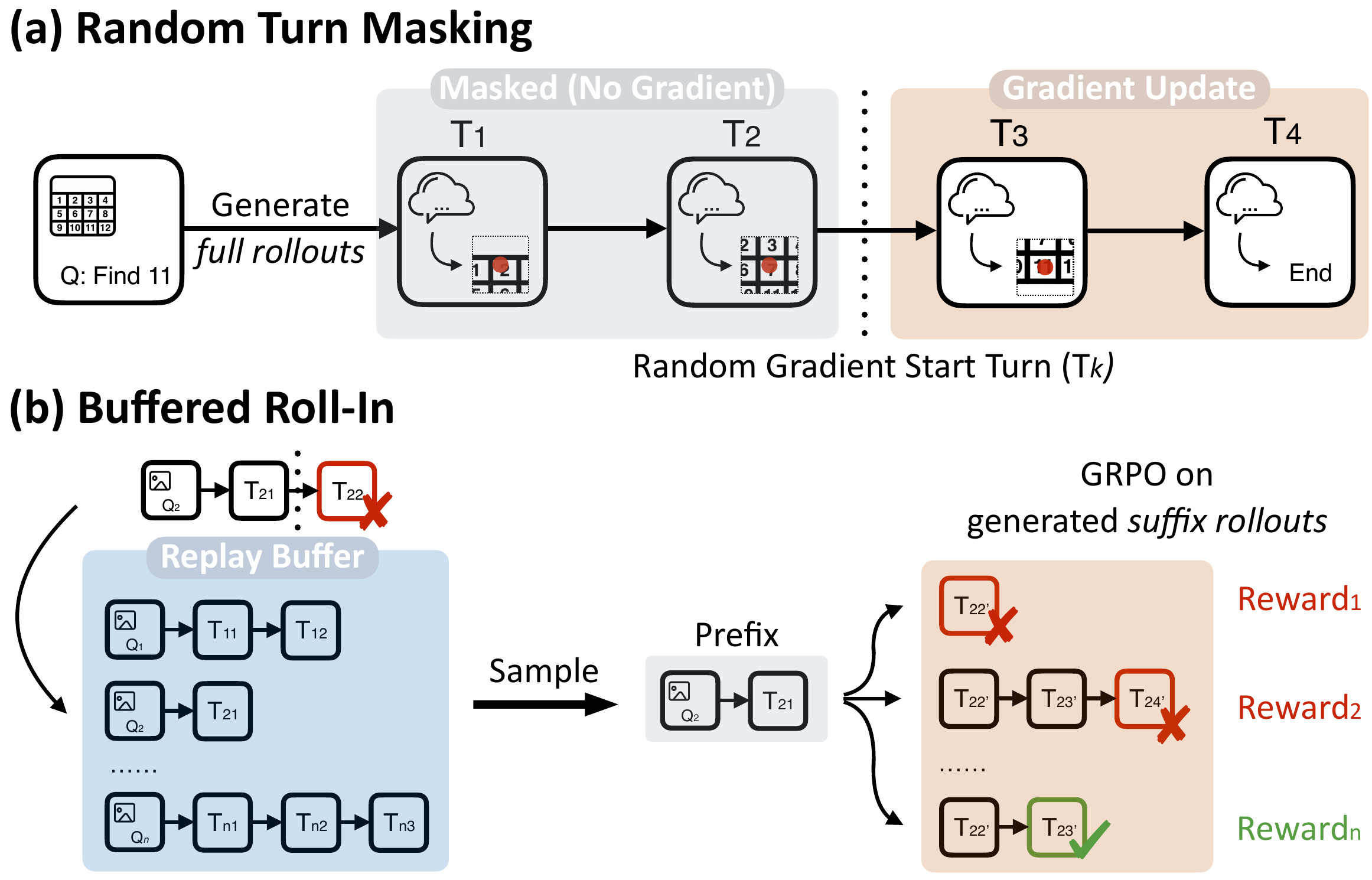}
    \caption{(a) \textbf{Random Turn Masking} masks gradient updates on a random number of prefix steps to avoid training on potentially erroneous steps. (b) \textbf{Buffered Roll-In} begins roll-outs from a potentially erroneous step in the replay buffer, enabling correction of this mistake. Note that partial rewards are assigned for correct reflection steps.}
    \label{fig:method}
\end{figure*}

\section{Problem Formulation: Multi-turn Inference with Reflection}
\label{sec:task_formulation}

\textbf{Reasoning with Multi-turn Inference.} We consider a multimodal reasoning task in which a model interacts with an environment $\mathcal{E}$. Given an input image $I$ and a natural language instruction $Q$, the goal is to answer $Q$ by taking a sequence of actions $a_t \in \mathcal{A}$ and receiving image observations $I_t \in \mathcal{O}$
% \gd{it bothers me a tiny bit that $i_t$ is lowercase and $I$ is capitalized but they're the same thing...not too big a deal though}
from the environment. A standard LVLM $\pi_\theta(a \mid I, Q)$ typically makes a one-shot decision by producing a single action and immediately terminating with a final answer. While simple, this single-turn setting does not allow the model to verify its prediction or recover from mistakes.

In this work, we formulate multimodal reasoning as a \emph{multi-turn sequential decision process}. Instead of relying on a single prediction, the model can iteratively propose an action, receive visual feedback from the environment, and decide whether to refine its prediction or terminate. Figure~\ref{fig:multi_vs_single} illustrates the difference between single-turn and multi-turn inference. We define this process by the tuple $(\mathcal{S}, \mathcal{A}, \mathcal{O}, \mathcal{R})$.

\textbf{State and Observation.} At turn $t$, the underlying state $s_t = (I, Q, \mathcal{H}_t) \in \mathcal{S}$
with the interaction history $\mathcal{H}_t = \{(a_0, I_0), \dots, (a_{t-1}, I_{t-1})\}$ for $a_{t} \in \mathcal{A}$. Upon taking an action $a_{t}$, the model receives an image $I_{t} \in \mathcal{O}$ as the visual observation  rendered by the environment, and can reflect on the observation to adjust its future actions. For example, in the visual grounding task in Figure~\ref{fig:multi_vs_single}, the model predicts a coordinate tuple for the given query. The environment then marks the predicted location on the image, e.g., with a red point, and returns a fixed-size crop centered at the predicted coordinates. This crop serves as the image observation $I_t$ for the next turn.

\textbf{Action Space.} An action $a_t \in \mathcal{A}$ corresponds to a reasoning step followed by an answer proposal or a termination: the policy $\pi_\theta(a_t \mid s_t)$ can propose an answer candidate that could be changed in later turns, or terminate the trajectory $\tau$ by finalizing an answer candidate from the last turn that it considers already correct. Upon termination, the final answer will be evaluated. In the visual grounding task, for example, an answer proposal at turn $t$ is a selection of pixel coordinates $p_{t} = (x_{t}, y_{t})$ within the image space, and a final answer needs to be locked in with a termination function call.

\textbf{Trajectory Reward.} The reward function $\mathcal{R}$ is only evaluated when the trajectory $\tau$ is terminated at turn $t=T$. The learning objective is to optimize the policy $\pi_\theta$ to maximize the final answer accuracy, enabling the model to refine its predictions iteratively based on visual feedback. Note that $\mathcal{R}$ involves multiple components that evaluate different aspects of the entire multi-turn trajectory, including final answer correctness, response format, and task progress at each turn (e.g., the reflection reward in Section~\ref{sec:our_rl_method}). 
\begin{figure*}
    \centering
    \includegraphics[width=\textwidth]{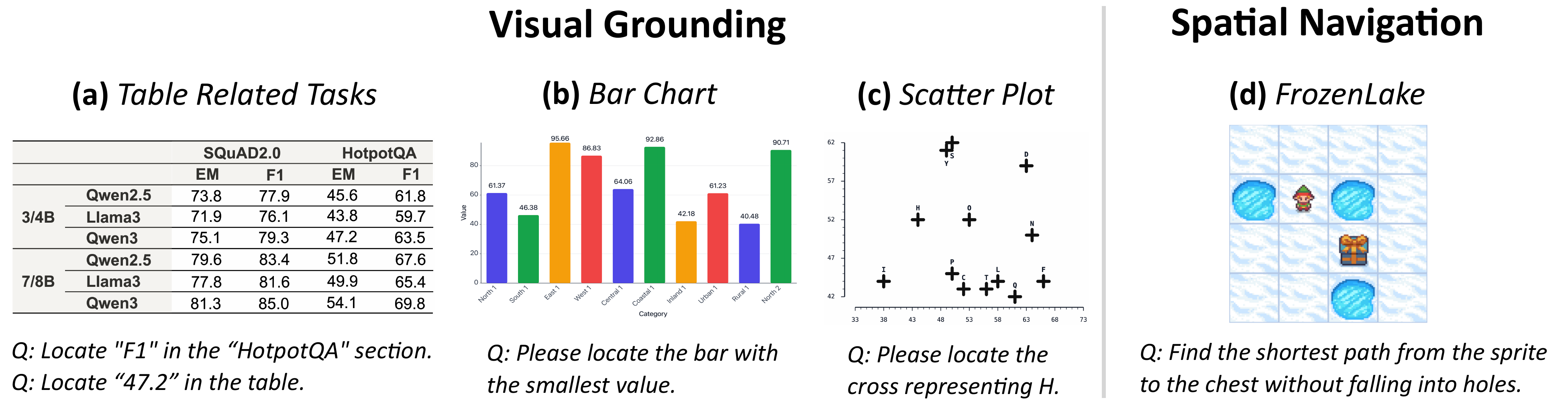}
    \caption{\textbf{Evaluation tasks for visually grounded self-reflection.} (1) \emph{Visual Grounding.}
    models are trained to localize headers of synthetic small tables. OOD tasks include generalization to larger tables (Large Table) queries of inner cell values rather than headers (Cell Query), and queries about bar charts \textbf{(b)}, and scatter plots \textbf{(c)}.  (2) \emph{Spatial Navigation.} The model is given a maze-style map and needs to find the shortest path from the start to the goal without running into obstacles. We train models on smaller maps and evaluate on larger maps as OOD tests.
    }
    \label{fig:ood_tasks}
\end{figure*}
\section{Methods}

\textbf{Training Framework.} Our training pipeline follows a two-stage \textbf{SFT $\rightarrow$ RL} recipe designed to instill and then refine reflective capabilities \citep{Guo2025,kimiteam2025kimik15scalingreinforcement, zhai2024finetuning, feizi2026grounding,sprague2026skillfactory}. Our goal is to improve the generalizability of the trained model over OOD tasks.

\subsection{Stage 1: Supervised Fine-Tuning (SFT)}

In the first stage, we first construct an offline dataset $\mathcal{D}_{\text{SFT}}$ with trajectories containing both immediate correct answers (single-turn) and iterative corrections (multi-turn): single-turn trajectories aim at teaching the model the visual reasoning task, and multi-turn trajectories teach the model the format of reflection based on visual feedback from the environment.

Each instance consists of an image $I$, an instruction $Q$, and a trajectory $\tau$. Single-turn trajectories consist of a single action leading to the correct answer, the visual feedback $I_0$, and a termination action $a_1$, $\tau = \{a_0, I_0, a_1\}$. A multi-turn trajectory $\tau = \{a_0, I_0, \dots, a_{T-1}, I_{T-1}, a_T\}$ consists of alternating visual feedback $I_t$ and assistant responses $a_t$ (CoT and answer candidates). Note that the last action $a_T$ is always a termination action.

On multi-turn trajectories, we optimize the standard auto-regressive cross-entropy loss over the assistant turns $a_{1},...,a_{T}$. Note that $a_{0}$ is excluded from the multi-turn loss update because $a_{0}$ is constructed to be erroneous so that the model needs to correct in the following assistant turns. This stage establishes the \emph{interaction format} and initializes the policy with basic error-correction behaviors.

\subsection{Stage 2: RL with Random Turn Masking and Buffered Roll-In}
\label{sec:our_rl_method}

To explicitly train the model to recover from errors, we do reinforcement learning on a set of examples $\mathcal{D}_\mathrm{RL}$. Our reward $\mathcal{R}$ is as defined below. We employ Group Relative Policy Optimization (GRPO; \citet{shao2024deepseekmath}) augmented with two novel mechanisms: \textbf{Random Turn Masking (RTM)} and \textbf{Buffered Roll-In} (see Figure~\ref{fig:method}).

\textbf{Reward Function.} We combine three reward components: format validity, answer correctness, and reflection shaping. (1) \emph{Format Reward:} We assign $r_{\text{fmt}}=1$ if the model response utilizes the valid tool-call
% \gd{it can be single-turn also right? I would just say valid format} 
format, and $0$ otherwise. Invalid formats result in a total trajectory reward of 0. (2) \emph{Outcome reward:} We assign $R_{\text{answer}}=1.0$ if the final answer is correct based on the task-specific metric, and $0$ otherwise. (3) \emph{Reflection reward:} To encourage convergence toward the target through multi-turn inference, we define a reflection reward  $R_{\text{refl}}$ based on an improvement-based reward function $\phi(\tau)$. Reflection reward is task-specific, and details for the evaluation tasks can be found in Section~\ref{sec:task_setup}; ablations in Section~\ref{sec:ablation} show that reflection reward elicits self-reflection better than the outcome reward.

The final scalar reward $R$ for a rollout is:

{\footnotesize

\begin{equation*}
R =
\begin{cases}

0 & \text{if } r_{\text{fmt}} = 0,\\[4pt]

\max\!\left(R_{\text{answer}}, R_{\text{refl}}\right) & \text{if } r_{\text{fmt}} = 1
\end{cases}
\label{eq:reward_func}
\end{equation*}
}
\normalsize

For incorrect predictions ($R_{\text{answer}}=0$), we provide partial credits based on improvements.

For a given question $q \in \mathcal{D}_\mathrm{RL}$, we sample a group of $G$ trajectories $\{\tau_1, \dots, \tau_G\}$ from the old policy $\pi_{\theta_{\text{old}}}$ and follow the standard GRPO training objective 
(Appendix~\ref{sec:method_details}).

\textbf{Random Turn Masking (RTM).} Given a full trajectory rollout $\tau$ of length $T$, we sample a start index $k \sim \mathrm{Unif}(\{1,\dots,T\})$ and compute policy-gradient updates only on the suffix from turn $k$ to $T$, masking the loss for earlier turns. Formally, let $\mathcal{J}(\theta)$ denote the RL objective. Under RTM, the gradient estimate is

{\footnotesize
\begin{equation*}
    \nabla \mathcal{J}_{\text{RTM}}(\theta) = \mathbb{E}_{\tau \sim \pi_\theta} \mathbb{E}_{k \sim U(1, T)} \left[ \sum_{t=k}^{T} \nabla \log \pi_\theta(a_t | s_t) \hat{A}_t \right],
\end{equation*}
}
\normalsize
where $\hat{A}_t$ is the GRPO advantage estimate. Although prefix turns before $k$ do not directly contribute to the gradient, they determine the conditioning state $s_k$. RTM therefore trains the policy to optimize returns from arbitrary intermediate states, including potentially erroneous ones, without learning to make the mistakes in the erroneous states. We interpret RTM as a form of \emph{reweighted per-decision policy gradient}; a derivation is provided in Appendix~\ref{sec:method_details}.

\textbf{Buffered Roll-In.} RTM reweights gradient updates across turns, but it still relies on on-policy rollouts. As the policy improves, failure states become less frequent, reducing the amount of training signal for error recovery. To maintain a diverse set of difficult recovery scenarios, we introduce \emph{Buffered Roll-In}. We maintain a replay buffer $\mathcal{B}$ of previously generated prefixes.
% \liyan{it also have correct states} 
During rollout generation on training examples from $\mathcal{D}_{\mathrm{RL}}$, if a trajectory terminates with an incorrect final answer, we treat the state immediately before termination as a valid but unresolved intermediate state. We remove the termination action and store the resulting prefix $\tau_{\mathrm{pre}}$ in $\mathcal{B}$. During training, instead of always generating rollouts from scratch, we also sample prefixes from $\mathcal{B}$. For each sampled prefix $\tau_{\mathrm{pre}}$, the policy generates a group of $G$ suffix completions $\{\tau_{\mathrm{suf}}^{(1)}, \dots, \tau_{\mathrm{suf}}^{(G)}\}$, and GRPO is applied only to these generated suffixes. This directly trains the model to recover from previously failed intermediate states.

To balance exploring new questions with addressing past failures, we construct each training batch by sampling questions from $\mathcal{D}_\mathrm{RL}$ for RTM with probability $\rho$ and prefixes from the buffer $\mathcal{B}$ for Buffered Roll-In with probability $1-\rho$. The final optimization objective is:
\begin{equation*}
    \mathcal{J}_{\text{Total}}(\theta) = \rho \mathcal{J}_{\text{RTM}}(\theta) + (1-\rho) \mathcal{J}_{\text{Buff}}(\theta)
\end{equation*}

This creates a self-paced curriculum: as the policy improves with RTM, the buffer naturally accumulates ``harder'' failure modes that the current policy struggles to resolve.

\section{Task Setup}
\label{sec:task_setup}

We study multi-turn inference with self-reflection in two visual feedback environments: visual grounding and spatial navigation. In both tasks, the model proposes an intermediate answer, receives an image observation $I_t$ from the environment, and either refines its prediction or terminates with a final answer.

\subsection{Visual Grounding}
\label{sec:table_setup}

Visual grounding \citep{kazemzadeh-etal-2014-referitgame, flickrentitiesijcv, Yu-Mattnet-2018, mao2016generation} aims to localize image regions referred to by natural language. We focus on data visualizations, where precise localization is challenging because tables and charts differ substantially from the real-world image distributions commonly seen during pre-training \citep{wang2024charxiv,tang2025chartmuseum}. This provides a controlled testbed for studying whether models can use visual feedback to self-correct under distribution shifts.

\textbf{Task and Environment.} Given an image $I$ and instruction $Q$, the model must predict the target coordinate $p=(x,y)$. At each turn $t$, it either proposes a coordinate $p_t$ or terminates with a final prediction. After each proposal, the environment returns visual feedback $I_t$: a $200\times200$ crop centered at $p_t$ with a red marker indicating the proposed location.

\textbf{Reward and Evaluation.}
The outcome reward is $1.0$ if the final coordinate $p_T$ falls within a Euclidean distance threshold $\delta_{\text{tol}}=40\,\mathrm{px}$ of the ground-truth coordinate, and $0$ otherwise. We report accuracy under this criterion, except for \emph{Bar Chart}, where a prediction is correct if it falls within the target bar's bounding box. We also use a distance-based reflection reward during training (Appendix~\ref{sec:method_details}).

\textbf{Data and Splits.}
We synthesize tables and charts following prior data generation procedures \citep{chartbench,zheng-etal-2025-tabledreamer}. The training and in-distribution test sets use small arXiv-style tables with row- or column-header queries. We evaluate OOD generalization on four 1K-example test sets: (1) \emph{Large Tables}, which increases table size; (2) \emph{Cell Query}, which asks for inner table cells; (3) \emph{Bar Chart}, which transfers grounding to bar charts; (4) and \emph{Scatter Plot}, which requires localizing labeled points. Figure~\ref{fig:ood_tasks} illustrates these tasks, with full generation details in Appendix~\ref{sec: sync_data_gen}.

\subsection{Spatial Navigation}
\label{sec:frozenlake_setup}

Spatial navigation is challenging for LVLMs \citep{wang2024is,stogiannidis2025mindgapbenchmarkingspatial} and provides a natural visual feedback environment because agent movement can be inspected and corrected over multiple turns. We use FrozenLake \citep{wu2025vsp, brockman2016openaigym}, a grid-based navigation task where the agent must reach a goal while avoiding holes and impassable obstacles. Prior work shows that textual chain-of-thought alone is insufficient for such tasks \citep{xu2026visual}.

\textbf{Task and Environment.} Given an input map image $I$ and a fixed instruction $Q$, the model predicts the shortest, valid path from the start to the goal that does not run into holes or walls. A path proposal is a sequence of actions $v_t=(v_t^1,v_t^2,\dots,v_t^L)$, where each action is in $\{\texttt{left}, \texttt{right}, \texttt{up}, \texttt{down}\}$. At each turn, the model either proposes a path or terminates with a final prediction. After each proposal, the environment returns visual feedback $I_t$ by drawing a segmented red line for the predicted path on the map. Figure~\ref{fig:example-visual-frozenlake-1} shows a detailed example.

\textbf{Reward and Evaluation.}
Following \citet{xu2026visual}, we use exact match (EM) as both the outcome reward and evaluation metric. A trajectory receives reward $1.0$ if the final path $v_T$ is valid and matches the optimal solution length, and $0$ otherwise. We additionally use an improvement-based reflection reward that measures partial progress toward the optimal path, detailed in Appendix~\ref{sec:method_details}.

\textbf{Data and Training.}
We adopt the training and evaluation data from \citet{xu2026visual}, using $3$--$5$ grid maps as the in-distribution training setting and larger $6\times6$ and $7\times7$ maps as OOD evaluations. Because spatial navigation is difficult for LVLMs, we first warm-start the model with direct-answer training on in-distribution examples without visual feedback, following \citet{xu2026visual}, and then apply all subsequent training methods to this warm-started model. Details can be found in Appendix~\ref{sec:frozenlake_data}.

\begin{table*}[t]
\centering
\small
\begin{tabular}{rccccc|c}
\toprule
\multicolumn{1}{l}{} & \multirow{3}{*}{\textbf{In-Distribution}} & \multicolumn{5}{c}{\textbf{Out-of-Distribution}} \\
\cmidrule(r){3-7}
\multicolumn{1}{l}{} &  & \textbf{\emph{Large Table}} & \textbf{\emph{Cell Query}} & \textbf{\emph{Bar Chart}} & \textbf{\emph{Scatter Plot}} & \textbf{\emph{OOD Avg}} \\
\midrule
\multicolumn{1}{l}{\emph{\textbf{Zero-shot}}} &  &  &  &  &  &  \\
$\text{Qwen2.5-VL-3B}_\text{Single}$      & 5.3  & 4.8  & 3.4  & 13.1     &   2.9   & 6.0 \\
$\text{Qwen2.5-VL-3B}_\text{Multi}^*$   & 5.6 (77.3\%)  & 2.2 (69.2\%) & 1.3 (86.1\%) &   12.0 (14.7\%)    &  2.4 (75.5\%)    & 4.5 \\
$\text{Qwen2.5-VL-7B}_\text{Single}$      & 17.9 & 9.7  & 6.1  &  15.1    &  4.5    & 8.8 \\
$\text{Qwen2.5-VL-7B}_\text{Multi}^*$   & 19.8 (58.0\%) & 8.9 (49.3\%) & 7.2 (76.8\%) &  14.7 (15.5\%) &   4.6 (38.2\%)   & 8.8 \\
VL-Rethinker-7B       & 15.1 & 7.9 & 5.7 &  1.0     &  3.7     & 4.6 \\
VL-Rethinker-32B       & 42.4 & 18.7 & 27.6  & 9.5     & 32.7     & 22.1 \\
\midrule
\multicolumn{7}{c}{\textit{Qwen2.5-VL-3B-Instruct}} \\
\midrule 
\multicolumn{1}{l}{\emph{\textbf{SFT}}} &  &  &  &  &  &  \\
$\text{Single-SFT}$            & 80.4 & 46.1 & 2.4 &  25.7    &  23.8    & 24.5 \\
$\text{Multi-SFT}$               & 84.7 & 50.4 & 1.6 &  13.1    &  24.8    & 22.5 \\
\text{Reflection Tuning}             & 92.7  &52.5  & 7.0  & 25.0      &   27.4 & 28.0 \\
\midrule
\multicolumn{1}{l}{\emph{\textbf{RL}}} &  &  &  &  &  &  \\
$\text{Single-SFT} \to \text{GRPO}$      & 96.2 & 53.3 & 5.3 &   27.1   &   34.7   & 30.1 \\
$\text{Multi-SFT} \to \text{GRPO}$        & 99.6 & 78.6 & 13.5 &   30.7   &   37.2   & 40.0 \\
\textbf{$\text{VRRL} (\text{Ours})$}                   & 99.6 & \textbf{88.6} & \textbf{20.3} & \textbf{33.5}     &  \textbf{40.3}    & \textbf{45.7} \\
\midrule
\multicolumn{7}{c}{\textit{Qwen2.5-VL-7B-Instruct}} \\
\midrule 
\multicolumn{1}{l}{\emph{\textbf{SFT}}} &  &  &  &  &  &  \\
$\text{Single-SFT}$            & 83.6 & 62.8 & 34.0  &20.3     & 68.9    & 46.5 \\
$\text{Multi-SFT}$               & 84.8 & 66.2 & 39.1  & 20.9   & 73.3    & 49.9 \\
\text{Reflection Tuning}             & 95.3  & 75.1  & 51.8   &  14.2     &  81.3  & 55.6 \\
\midrule
\multicolumn{1}{l}{\emph{\textbf{RL}}} &  &  &  &  &  &  \\
$\text{Single-SFT} \to \text{GRPO}$        & 99.6  & 89.6 & 68.3 &  38.6    & 84.3     & 70.2 \\
$\text{Multi-SFT} \to \text{GRPO}$        & 99.6 &  \textbf{91.4} & 68.4 & 46.8      & 86.0  & 73.2 \\
\textbf{$\text{VRRL} (\text{Ours})$}                   &  99.7 & 89.6  & \textbf{77.3} & \textbf{57.0}     &  \textbf{89.7}  & \textbf{78.4} \\
\bottomrule
\end{tabular}
\caption{\textbf{In-distribution and OOD evaluation results for visual grounding across models}. We perform paired bootstrap tests to compare the best-performing model with the second-best model in each column. Bold indicates that the best result is better by a statistically significant margin ($p < 0.05$). For $\text{Qwen2.5-VL-3B}_\text{Multi}$ and $\text{Qwen2.5-VL-7B}_\text{Multi}$, we report the percentage of traces where the model's reflection turns repeat the same predictions as previous turns.} \label{tab:main_results}
\end{table*}

\section{Experimental Setup}

We compare our proposed method against baseline methods below. Implementation details and example outputs can be found in Appendix~\ref{sec:implementation_details} and~\ref{sec:example_outputs}.

\textbf{Zero-Shot Baselines.} We evaluate \textbf{Qwen2.5-VL-3B-Instruct} and \textbf{7B} models \citep{Qwen2.5-VL} under two settings: direct pointing and a multi-turn reflection prompt that asks the model to critique and refine its coordinates without parameter updates. For spatial navigation tasks, we also evaluate \textbf{Qwen3-VL-4B-Instruct} \citep{bai2025qwen3vltechnicalreport}.

\textbf{Supervised Fine-Tuning (SFT) Baselines.} We compare two SFT strategies for Qwen2.5-VL-3B/7B on visual grounding, and Qwen2.5-VL-3B and Qwen3-VL-4B on spatial navigation. \textbf{$\text{Single-SFT}$} trains on perfect single-turn trajectories, where the model outputs the correct coordinate and terminates immediately. \textbf{$\text{Multi-SFT}$} trains on a mixture of perfect and synthetic recovery trajectories, where the model first predicts an incorrect coordinate and then iteratively uses visual feedback to refine its prediction until correct.

\begin{figure*}[t]
    \centering
    \includegraphics[width=\linewidth]{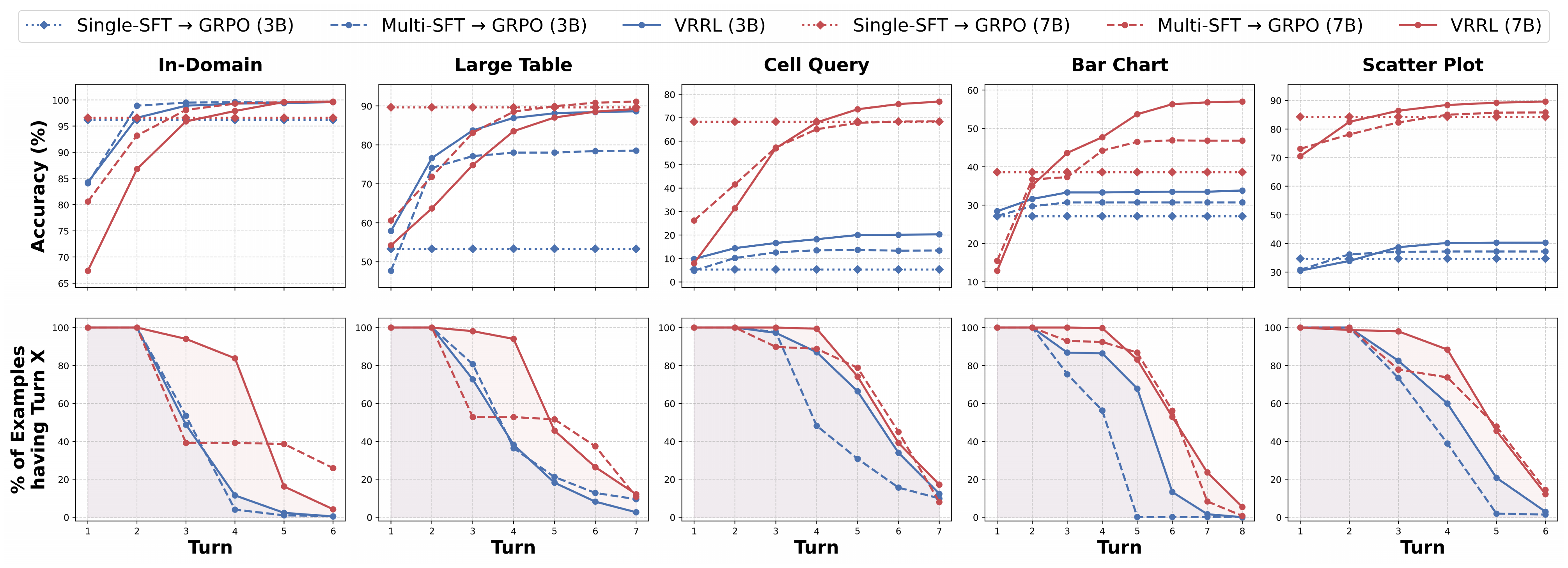}
    \caption{\textbf{Performance for multi-turn reflection across in-distribution and OOD tasks for visual grounding tasks.} The top row illustrates the progression of cumulative accuracy across turns. $\text{Single-SFT} \to \text{GRPO}$ is a single-turn baseline method where its per-turn accuracy remains the same. The bottom row shows the percentage of examples reaching turn X. VRRL generally uses reflection better, improving more over iterations.
    }
    \label{fig:reflection_turn}
\end{figure*}

\textbf{RL Baselines.} We apply standard GRPO to $\text{Single-SFT}$ and $\text{Multi-SFT}$ (\textbf{$\text{Single/Multi-SFT} \to \text{GRPO}$}). These baselines (1) use a sparse binary outcome reward (1 for success, 0 for failure) without the reflection reward shaping $R_{\text{refl}}$; (2) calculate loss over the full trajectory without Random Turn Masking; and (3) do not use Buffered Roll-In.

\textbf{Reflection-Oriented Baselines.} We include prior fine-tuning approaches designed to instill self-reflection in LVLMs as baselines: (1) \textbf{VL-Rethinker} \citep{wang2026vlrethinker} augments GRPO training with selective sampling and forced rethinking to encourage self-reflection \emph{through textual reasoning only} by scaling long CoT traces. This differs from our setting, which focuses on visually grounded self-reflection driven by visual feedback. We directly evaluate the released VL-Rethinker models trained on diverse multimodal reasoning tasks for their OOD generalization capabilities. (2) \textbf{Reflection Tuning} \citep{wu2025guireflection} trains LVLMs to perform self-reflection based on visual feedback through iterative online SFT, where the model learns from corrections of its own generated error trajectories. Since this approach is naturally compatible with our visually grounded setting, we apply reflection tuning on top of our multi-turn SFT model using the same in-distribution training data.

\textbf{\textbf{VRRL} (Ours).} Our method applies our full training recipe on top of $\text{Multi-SFT}$.

\textbf{Training.} For visual grounding, we use 15K training examples for SFT and 6K for RL. For spatial navigation, we use 4K training examples for SFT and 2K for RL. Implementation details, dataset details, and training configurations can be found in Appendix~\ref{sec:implementation_details}.

\begin{table*}
\centering
\footnotesize
\renewcommand{\tabcolsep}{1.0mm}
\begin{tabular}{lccc|c||ccc|c}
\toprule
& \multicolumn{4}{c||}{\textit{Qwen2.5-VL-3B-Instruct}} & \multicolumn{4}{c}{\textit{Qwen3-VL-4B-Instruct}} \\
\cmidrule(lr){2-5}\cmidrule(lr){6-9}
& \textbf{ID} & \multicolumn{3}{c||}{\textbf{OOD}} & \textbf{ID} & \multicolumn{3}{c}{\textbf{OOD}} \\
\cmidrule(lr){3-5}\cmidrule(lr){7-9}
& \textbf{Avg.} & \textbf{$6\times6$} & \textbf{$7\times7$} & \textbf{Avg.}
& \textbf{Avg.} & \textbf{$6\times6$} & \textbf{$7\times7$} & \textbf{Avg.} \\
\midrule
\midrule
\multicolumn{9}{l}{\emph{\textbf{Zero-shot}}} \\
Single             & 2.4  & 1.2  & 1.6  & 1.4  & 8.4  & 2.4  & 2.0  & 2.2  \\
Multi              & 2.8  & 0.8  & 2.0  & 1.4  & 34.0 & 8.8 & 2.4 & 5.6 \\
VL-Rethinker-7B    & 11.3 & 3.2  & 3.2  & 3.2  & -    & -    & -    & -    \\
VL-Rethinker-32B   & 26.5 & 10.8 & 3.6  & 7.2  & -    & -    & -    & -    \\
\midrule
\multicolumn{9}{l}{\emph{\textbf{SFT}}} \\
$\textit{Base}^{*}$ & 77.7 & 33.2 & 4.8  & 19.0 & 88.1 & 49.6 & 5.6  & 27.6 \\
Single-SFT          & 83.5 & 39.2 & 8.8  & 24.0 & 90.7 & 56.8 & 8.0  & 32.4 \\
Multi-SFT           & 81.2 & 41.6 & 10.2 & 25.9 & 86.9 & 60.4 & 21.2 & 40.8 \\
Reflection Tuning   & 85.7 & 49.6 & 12.4 & 31.0 & 93.6 & 63.6 & 13.2 & 38.4 \\
\midrule
\multicolumn{9}{l}{\emph{\textbf{RL}}} \\
$\text{Single-SFT} \to \text{GRPO}$ & 85.9 & 49.2 & 8.8 & 29.0 & 93.2 & 62.4 & 5.6  & 34.0 \\
$\text{Multi-SFT} \to \text{GRPO}$  & 85.2 & 42.8 & 9.2 & 26.0 & 87.9 & 61.6 & 10.4 & 36.0 \\
\textbf{VRRL (Ours)} & 83.9 & \textbf{54.8} & \textbf{23.6} & \textbf{39.2}
                     & 89.1 & 65.2 & \textbf{39.2} & \textbf{52.2} \\
\bottomrule
\end{tabular}
\caption{\textbf{In-distribution (ID) and OOD evaluation for spatial navigation on Qwen2.5-VL-3B-Instruct and Qwen3-VL-4B-Instruct}.
\emph{Base}$^{*}$: We first train models on ID data with direct answering without visual feedback to obtain \emph{Base} as a warm-started model on the task distribution, and then apply all training methods based upon this model. We perform paired bootstrap tests to compare the best-performing model with the second-best model in each column. Bold indicates that the best result is statistically significant ($p < 0.05$). }
\label{tab:frozenlake_results}
\end{table*}
\section{Results}
\label{sec:results}
\subsection{Visual Grounding}
\label{sec:table_results}
\textbf{Prompting does not elicit reliable self-reflection.}
Table~\ref{tab:main_results} shows that off-the-shelf LVLMs struggle with precise spatial localization. In the zero-shot setting, both 3B and 7B models achieve low accuracy across tasks: Prompting models to reflect provides little benefit and can even hurt performance. In many multi-turn traces, zero-shot models simply repeat previous predictions without meaningful correction. 
 
Furthermore, VL-Rethinker models that have been trained to reflect with textual CoT reasoning underperform on OOD tasks; their reflective CoT traces generally fail to correct mistakes. These results suggest that visually grounded self-correction cannot be reliably elicited through prompting or textual CoT alone \citep{wu2026better, wu2025visco, huang2025highresolutionvisualreasoningmultiturn, jiang2025mmecot}. 

\textbf{SFT is limited.} SFT methods, including  $\text{Multi-SFT}$ and reflection tuning that specifically instill visual self-reflection, mainly  teach the in-distribution knowledge, but remain brittle on the evaluated OOD tasks. For example, $\text{Multi-SFT}$ only learns the format of reflection rather than robust error-correcting behavior.

\textbf{RL on multi-turn reflection models improves OOD generalization.} 
RL substantially improves OOD performance over SFT. Even single-turn RL, $\text{Single-SFT} \to \text{GRPO}$, raises \emph{Large Table} accuracy to 53.3\% for 3B and 89.6\% for 7B. However, the main gains come from combining RL with a multi-turn reflection formulation: $\text{Multi-SFT} \to \text{GRPO}$ improves the 3B model to 78.6\% on \emph{Large Table}, 13.5\% on \emph{Cell Query}, 30.7\% on \emph{Bar Chart}, and 37.2\% on \emph{Scatter Plot}, yielding 3--25\% absolute gains over its single-turn counterpart. VRRL further improves over $\text{Multi-SFT} \to \text{GRPO}$ by 3--10\% on most OOD tasks while maintaining near-perfect in-distribution accuracy for both model scales. These gains are notable because the OOD splits require generalization across table size, query type, and visual domain, indicating that VRRL induces a more robust visual grounding capability.

\textbf{VRRL teaches effective reflection.}
Figure~\ref{fig:reflection_turn} shows that VRRL's gains come from improved multi-turn correction rather than stronger one-shot grounding alone. While $\text{Single-SFT} \to \text{GRPO}$ improves single-turn accuracy, it lacks the turn-by-turn refinement behavior of multi-turn RL. $\text{Multi-SFT} \to \text{GRPO}$ improves across early reflection turns, but VRRL converts this behavior into stronger OOD generalization. The turn-distribution plots further show that VRRL adapts its number of refinement steps to task familiarity: it terminates early on in-distribution examples, but continues refining on OOD tasks and achieves higher accuracy in later turns.

\begin{table}[]
\centering
\scriptsize
\begin{tabular}{lrrrr}
\toprule
\multirow{2}{*}{}  & \multicolumn{2}{c}{\textbf{ID}}                  & \multicolumn{2}{c}{\textbf{OOD}}                         \\
\cmidrule(r){2-3} \cmidrule(r){4-5}
                   & \multicolumn{1}{l}{\# Turns} & $\Delta_\mathrm{ref}$ & \multicolumn{1}{l}{\# Turns} & $\Delta_\mathrm{ref}$         \\
                   \midrule
                   \midrule
                   \multicolumn{5}{c}{\textit{Qwen2.5-VL-3B-Instruct}} \\
                   \midrule
                   \midrule
$\text{Single-SFT} \to \text{GRPO}$  & 2.00                           & + 0    & 2.00                           & + 0            \\
$\text{Multi-SFT}$     & 2.41                           & + 0.1  & 3.51                           & + 2.8          \\
$\text{Multi-SFT} \to \text{GRPO}$  & 2.00                           & + 0    & 2.00                           & + 0            \\
Reflection Tuning  & 2.23                           & + 1.3  & 3.17                           & + 3.4          \\
\textbf{$\text{VRRL} (\text{Ours})$}              & 2.37                           & + \textbf{6.0}  & 3.39                           & \textbf{+ 13.6} \\
\midrule
                   \midrule
\multicolumn{5}{c}{\textit{Qwen3-VL-4B-Instruct}} \\
                   \midrule
                   \midrule
$\text{Single-SFT} \to \text{GRPO}$  & 2.00                           & + 0    & 2.00                           & + 0            \\
$\text{Multi-SFT}$     & 2.24                           & + 1.1  & 4.57                           & + 11.0          \\
$\text{Multi-SFT} \to \text{GRPO}$  & 2.00                           & + 0    & 2.00                           & + 0            \\
Reflection Tuning  & 2.24                           & + 1.7  & 4.69                           & + 7.4          \\
\textbf{$\text{VRRL} (\text{Ours})$}              & 2.22                           & + 1.1  & 4.64                           & \textbf{+ 23.0} \\
\bottomrule
\end{tabular}

\caption{\textbf{Reflection behaviors of training methods for spatial navigation}. \emph{\# Turns} denotes the average number of turns, including a termination turn, of the model responses, and \emph{$\Delta_{\mathrm{ref}}$} is the improvement of task accuracy by multi-turn reflection inference.}
\label{tab:frozenlake_analysis}
\end{table}
\subsection{Spatial Navigation}
\label{sec:frozenlake_results}

\textbf{VRRL improves OOD spatial navigation.} Table~\ref{tab:frozenlake_results} shows results on our spatial navigation task. VRRL achieves comparable in-distribution accuracy to the baselines while outperforming all baselines on OOD settings for both models. On OOD tasks, VRRL improves multi-turn SFT by 13.3\% on Qwen2.5-VL-3B and 11.4\% on Qwen3-VL-4B, and it outperforms the second-best baselines by 9.8 \% on average across two models.

\textbf{VRRL uses reflection more efficiently.}
We next analyze reflection behavior in spatial navigation. Table~\ref{tab:frozenlake_analysis} reports the average number of turns, including the required termination call, as well as the performance improvement from multi-turn reflection. Standard GRPO with only outcome reward ($\text{Multi-SFT} \to \text{GRPO}$) largely suppresses the reflection behavior learned during multi-turn SFT in this setting. In contrast, VRRL uses reflection more effectively and selectively: it improves performance over turns while requiring similar number of turns on average than other reflection-oriented baselines.

\section{Ablations}

\label{sec:ablation}

\begin{table*}[]
\centering
\small
\begin{tabular}{lcccc|c}
\toprule
                                              & \multicolumn{5}{c}{\textbf{Out-of-Distribution}} \\
\cmidrule(r){2-6}
                                              & \textbf{\emph{Large Tabl}e} & \textbf{\emph{Cell Query}} & \textbf{\emph{Bar Chart}} & \textbf{\emph{Scatter Plot}} & \textbf{Average} \\
\midrule
$\text{Single-SFT} \to \text{GRPO}$                  & 53.3               & 5.3  & 27.1 & 34.7 & 30.1 \\
$\text{Multi-SFT} \to \text{GRPO}$                 & 78.6               & 13.5  & 30.7 & 37.2 & 40.0 \\
\midrule
\multicolumn{1}{r}{+ RTM}                     & 70.5               & 11.4  & 33.8 & 34.1 & 37.5 \\
\multicolumn{1}{r}{+ Buffered Roll-In}        & 63.2               & \textbf{26.1} & \textbf{43.9}  & 37.1 & 42.6 \\
\multicolumn{1}{r}{+ RTM + Buffered Roll-In}  & 78.6               & 18.1  & 36.9 & 39.9 & 43.4 \\
\multicolumn{1}{r}{+ Reflection Reward}       & 79.1   & 19.5  & 38.0 & 34.1 & 42.7 \\
\midrule
VRRL (Ours)
                                              & \textbf{88.6}      & 20.3 & 
                                              33.5 & \textbf{40.3} & \textbf{45.7} \\
\bottomrule
\end{tabular}
\caption{Ablation study on VRRL over the 3B model (Reflection Reward + RTM + Buffered Roll-In) on visual grounding. We find that with only the Buffered Roll-In method, the model fails to learn multi-turn reflection and terminates immediately after one pointing call. However, it still outperforms the single-turn model $\text{Single-SFT} \to \text{GRPO}$  on all settings by a large margin.}
\label{tab:ablation}
\end{table*}

\begin{table}
\centering
\footnotesize
\renewcommand{\tabcolsep}{1.0mm}
\begin{tabular}{lccc|c}
\toprule
\multirow{2}{*}{}  & \multicolumn{1}{c}{\textbf{ID}}      & \multicolumn{3}{c}{\textbf{OOD}}                                                                 \\
\cmidrule(r){3-5}
                   & \multicolumn{1}{c}{\textbf{Avg}} & \multicolumn{1}{c}{\textbf{$6\times6$}} & \multicolumn{1}{c}\textbf{{$7\times7$}} & \multicolumn{1}{c}{\textbf{Avg}} \\
               
                   \midrule
                   \midrule
                   \multicolumn{5}{c}{\textit{Qwen2.5-VL-3B-Instruct}} \\
                   \midrule
                   \midrule

$\text{Single-SFT} \to \text{GRPO}$      & 85.9                        & 49.2                        & 8.8                         & 29.0                        \\
$\text{Multi-SFT} \to \text{GRPO}$    & 85.2                        & 42.8                        & 9.2                         & 26.0                        \\
\midrule
+ RTM    & 84.4                        & 48.0              & 9.2               & 28.6 \\
+ Buffered Roll-In   & 85.3                        & 47.6               & 7.6               & 27.6 \\
+ RTM + Buffered Roll-In    & 83.3                        & 52.4               & 13.2               & 32.8 \\
+ Reflection Reward    & 86.9                        & 51.6              & 10.0              & 30.8 \\
\midrule

\textbf{VRRL (Ours)}    & 83.9                        & \textbf{54.8}               & \textbf{23.6}               & \textbf{39.2} \\
\bottomrule
\end{tabular}

\caption{Ablation study on VRRL over the 3B model (Reflection Reward + RTM + Buffered Roll-In) on spatial navigation.  }
\label{tab:frozenlake_ablation}
\end{table}

Table~\ref{tab:ablation} and~\ref{tab:frozenlake_ablation} show ablation studies for visual grounding and spatial navigation, respectively, to isolate the impact of each proposed component. When added individually to the $\text{Multi-SFT} \to \text{GRPO}$ baseline, RTM and Buffered roll-in produce mixed results. In particular, buffered roll-in alone causes the model to collapse into mostly single-turn behavior, losing the reflection capability learned during multi-turn SFT. However, by exposing the model to diverse intermediate states, it still improves average single-turn performance, leading to 42.6\% on visual grounding and substantially outperforming $\text{Single-SFT} \to \text{GRPO}$ at 30.1\%.

Importantly, combining RTM with buffered roll-in restores and enhances multi-turn behavior. This combination resolves the single-turn collapse, restoring \emph{Large Table} performance to the baseline level and improving all other OOD settings, demonstrating that these two components are highly complementary. Reflection reward is also necessary to achieve strong performance for both tasks, indicating that the best OOD robustness is obtained only when all three components are combined.
Furthermore, the reflection reward alone proves to be an effective individual addition (for example, 42.7\% average on visual grounding). By providing a reward shaping signal rather than a sparse binary reward, it teaches the model to iteratively move toward better states even when the final answer remains incorrect, which is particularly useful when the query type is unseen. The full model outperforms all ablated baselines, indicating that the best OOD robustness is obtained only when all three components are combined.

\section{Related Work}

\textbf{Self-Reflection of LVLMs.} Self-reflection, the process of inspecting previously generated outputs and correcting potential mistakes, has emerged as an important capability of LLMs \citep{madaan2023selfrefine, gou2024critic} and has been applied to a variety of downstream tasks \citep{shinn2023reflexion,wadhwa-etal-2024-learning-refine}. As pre-trained LVLMs only exhibit limited self-reflection capabilities \citep{cheng-etal-2025-vision}, post-training methods have been explored to further improve this behavior \citep{li2025revisor, huang2024large}. 
As existing approaches mainly train LVLMs for self-reflection by scaling CoT reasoning in text \citep{wan2025srpo, yang2025look, chung2025v1learningpointvisual,jian-etal-2025-look,wang2026vlrethinker}, they often over-rely on the textual information in prompts \citep{vo2026vision,tang2026scaling}, and fail to utilize visual feedback to adjust predictions. In contrast, \emph{visually grounded self-reflection}, where models learn self-verification and error recovery by leveraging image inputs and intermediate visual feedback, is less explored.

\textbf{Thinking with Images.} A recent line of work studies the paradigm of ``thinking with images'' \citep{su2025thinking}, in which VLMs are augmented with external tools (e.g., OCR, depth analysis, zooming, and image segmentation) that provide new visual information as tool outputs across multiple turns \citep{su2025openthinkimg, hu2024visual, shao2024visual}. These methods equip VLMs with the ability to make accurate visual tool calls \citep{wu2026vtoolr} and to integrate tool outputs into their reasoning process \citep{huang2025highresolutionvisualreasoningmultiturn, wang2026vgr, yang2025thinking}, but they mainly focus on incentivizing correct tool calls from VLMs rather than reflective capability of VLMs over tool outputs to validate their candidate answers.

\section{Conclusion} In this paper, we showed that multi-turn reflection can improve robustness of visual grounding and spatial navigation for LVLMs. To this end, we propose VRRL, an RL training method that combines Random Turn Masking and Buffered Roll-In to teach models both when to stop and how to recover from intermediate mistakes using visual feedback. Our method achieves strong in-distribution performance while substantially improving OOD generalization over zero-shot, SFT, and standard RL baselines. Our model successfully uses multiple steps of refinement to achieve this performance. Our work highlights multi-turn self-reflection as a key direction for improving the robustness and generalization of LVLMs.

\section*{Limitations}

We acknowledge several limitations in our work. First, our evaluation assumes visual feedback is available from the environment, which is not readily applicable to some multi-modal reasoning tasks such as open-ended VQA on real-world images. While this design choice was intentional, allowing us to isolate multi-turn grounding and refinement mechanics, it may not fully capture the complexity and diversity of real-world tasks.

Second, due to computational constraints, all training experiments are conducted with a single model family of Qwen (Qwen2.5-VL and Qwen3-VL), up to 7B model scale. While our results have clearly demonstrated the effectiveness of our training method, we have not evaluated whether the same conclusions hold consistently across larger model scale or different architectures.

\section*{Acknowledgments}
This work was supported by NSF CAREER Award IIS-2145280, NSF grant IIS-2433071, the NSF AI Institute for Foundations of Machine Learning (IFML), the NSF under Cooperative Agreement 2421782 and the Simons Foundation grant MPS-AI-00010515 awarded to the NSF-Simons AI Institute for Cosmic Origins — CosmicAI, \url{https://www.cosmicai.org/}, and an award from ExxonMobil. This work was also partially supported by the Sloan Foundation. Finally, this work has been supported by a compute grant from NVIDIA. We also acknowledge use of the research computing resources of the Empire AI Consortium, Inc., with support from the State of New York, the Simons Foundation, and the Secunda Family Foundation.

\bibliography{custom}

\appendix
\section{Method Details} \label{sec:method_details}

\paragraph{GRPO.}
For a given question $q$, we sample a group of $G$ trajectories
$\{\tau_1, \dots, \tau_G\}$ from the old policy
$\pi_{\theta_{\mathrm{old}}}$ and optimize the standard GRPO objective:

\begin{equation*}
\mathcal{J}_{\mathrm{GRPO}}(\theta)
=
\mathbb{E}_{q,\{\tau_i\}_{i=1}^G}
\left[
\frac{1}{G}
\sum_{i=1}^{G}
\frac{1}{|\tau_i|}
\sum_{t=1}^{|\tau_i|}
\ell_{i,t}(\theta)
\right],
\end{equation*}

where $\{\tau_i\}_{i=1}^G \sim \pi_{\theta_{\mathrm{old}}}$ and

\begin{equation*}
\ell_{i,t}(\theta)
=
\mathcal{L}^{\mathrm{clip}}_{i,t}(\theta)
-
\beta
\mathbb{D}_{\mathrm{KL}}
\left(
\pi_\theta \,\|\, \pi_{\mathrm{ref}}
\right).
\end{equation*}

The clipped objective is defined as
\begin{equation*}
r_{i,t}(\theta)
=
\frac{
\pi_\theta(a_{i,t} \mid s_{i,t})
}{
\pi_{\theta_{\mathrm{old}}}(a_{i,t} \mid s_{i,t})
},
\end{equation*}
{\scriptsize
\begin{equation*}
\mathcal{L}^{\mathrm{clip}}_{i,t}(\theta)
=
\min\left(
r_{i,t}(\theta) A_i,\,
\operatorname{clip}
\left(
r_{i,t}(\theta), 1-\epsilon, 1+\epsilon
\right) A_i
\right).
\end{equation*}
}

$\mathbb{D}_{\text{KL}}$ is the regularization term ensuring the policy does not deviate too much from the reference model and ${A}_{i}$ is the advantage computed by normalizing the trajectory reward within the group:

\begin{equation*}
    A_i = \frac{r_i - \text{mean}(\{r_1, \dots, r_G\})}{\text{std}(\{r_1, \dots, r_G\})}.
\end{equation*}

\textbf{Random Turn Masking (RTM) Interpretation.} We can interpret RTM as a form of \emph{reweighted per-decision policy gradient}. By expanding the expectation over the uniform starting index $k$, we observe that the contribution of each turn $t$ to the total gradient is weighted by the probability of it being included in the suffix. Let $w_t$ be the weight assigned to the gradient at turn $t$. Since a turn $t$ is included in the loss whenever the sampled start index $k \le t$, and $k$ is sampled uniformly, the effective weight is:

\begin{equation*}
    w_t = P(k \le t) = \sum_{j=1}^{t} \frac{1}{T} = \frac{t}{T}
\end{equation*}

Substituting this back into the gradient formulation, the expected RTM gradient is equivalent to a weighted standard policy gradient:

\begin{equation*}
     \mathbb{E} \left[ \nabla \mathcal{J}_{\text{RTM}}(\theta) \right] = \mathbb{E}_{\tau \sim \pi_\theta} \left[ \sum_{t=1}^{T} \frac{t}{T} \cdot \nabla \log \pi_\theta(a_t | s_t) \hat{A}_t \right]
\end{equation*}

This suggests that RTM applies a linear weighting schedule $w_t \propto t$. Later turns, which correspond to refinement and reflection steps, receive higher gradient magnitude than early turns. This implicitly prioritizes the optimization of \emph{recovery} behavior over initial exploration.

\textbf{Reward Function for Visual Grounding.} In this work, we define potential function $\phi(d)$ based on the Euclidean distance $d$ to the target:
\begin{equation*}
\phi(d) = \frac{1}{2}\left(\exp\left(-\frac{d^2}{\sigma_1^2}\right) + \exp\left(-\frac{d^2}{\sigma_2^2}\right)\right).
\end{equation*}

We select different $\sigma$ values to ensure the model receives meaningful feedback signals when the prediction is both far from (via $\sigma_2$) and near (via $\sigma_1$) the target. We choose to use improvement based reward function as we find that when the reflection reward is only based on the distance of final prediction to the ground truth, then the model will improve its prediction in the first turn and hence lose the ability to keep the reflection behavior. There is a large space in the design of reflection rewards and finding the best reflection reward is beyond the scope of this work as the best reflection reward may be task dependent.

We define the raw, unshaped improvement-based reward $r_{\text{refl}}$ based on $\phi(d)$. We define the following shaping to obtain the final reflection reward $R_{\text{refl}}$.
%{\footnotesize
\begin{equation*}
R_{\text{refl}} = 0.1 + 0.9 \cdot \max\!\left(0, r_{\text{refl}}\right) \  \text{if } r_{\text{fmt}} = 1
\end{equation*}
%}
%\normalsize
We give a weight of $0.9$ to the reflection reward and $0.1$ indicates the format reward when the format of the output is correct $r_{\text{fmt}} = 1$. For correct predictions, note that $r_{\text{refl}}$ is capped to $1.0$ given the definition of $\phi(d)$, so the total reward $R$ is capped to $1.0$ in combination with the format reward. For incorrect predictions, we provide partial credit based on improvement, using $\max(0,r_{\text{refl}})$ to avoid over-penalizing regressive reflection attempts: empirically, allowing negative shaping reduced the model's tendency to engage in reflection, since it can always achieve the base format reward (0.1) without attempting corrective moves.

\textbf{Reward Function for Spatial Navigation.} Following \citet{xu2026visual}, we build reflection reward upon \textbf{progress rate} for the spatial navigation task as follows. Given a model predicted path $\hat{v}$ and $M$ ground truth optimal paths $v_{m}$ for $m\in \{1,...,M\}$, the progress rate is computed as
\[
\mathrm{PR}
=
\max_{m\in\{1,\ldots,M\}}
\frac{1}{n}
\sum_{j=1}^{n}
\left[
\prod_{k=1}^{j}
\mathbb{I}
\left(
\hat{v}_k = v_k^{(m)}
\right)
\right]
\]
In other words, progress rate measures the ratio of consecutive correct steps from the start that align with at least one ground truth trajectory.

Then, the unshaped reflection reward is 

\[
r_{\text{refl}} = \operatorname{clip}\left(\displaystyle\sum_{t=2}^{T} w(\Delta \mathrm{PR}_t),\ 0,\ 1\right)
\]

  where $T$ is the total number of turns, $\Delta \mathrm{PR}_t = \mathrm{PR}_t - \mathrm{PR}_{t-1}$, and 
\[
  w(\Delta\mathrm{PR}) =
  \begin{cases}
  \Delta\mathrm{PR} & \text{if } \Delta\mathrm{PR} \geq 0 \\
  \lambda_{\text{deg}} \cdot \Delta\mathrm{PR} & \text{if } \Delta\mathrm{PR} < 0
  \end{cases}
  \]
where $\lambda_{\text{deg}}$ is a hyperparameter.

In other words, the design rewards improvement towards the optimal trajectories over multiple turns of refinement, and penalize unhelpful revisions that lead to regression in progress.

We define the following shaping to obtain the final reflection reward $R_{\text{refl}}$ when the response has the correct format (i.e., $r_{\text{fmt}} = 1$).

{\scriptsize

\begin{equation*}
R_\text{refl} =
\begin{cases}

0.1 + 0.9 \cdot \max\!\left(0, r_{\text{refl}}\right) + \alpha r_{\text{refl}} - \gamma T & \text{if } r_{\text{coord}} = 1,\\[4pt]
 0.1 + 0.9 \cdot \max\!\left(0, r_{\text{refl}}\right)  - \gamma T\ & \text{if }  r_{\text{coord}} = 0

\end{cases}
\end{equation*}
}
\normalsize
Where $r_{\text{coord}}$ is the correctness of the final answer, $\alpha$ is the reflection bonus coefficient, and $\gamma$ is the step cost penalty coefficient. 

The first two terms (format reward and weighted raw reflection reward) are the same as the reflection shaping for visual grounding, and we added the last two terms for spatial navigation specifically: (1) Reflection bonus: if the final answer is correct, we provide the model with a reward bonus of $\alpha$ if the response uses reflections. This provides incentive for the model to perform more multi-turn reflections during training for this difficult task; (2) Step cost: to prevent the model from over-reflecting to hack the reflection bonus, we apply a step cost of $\gamma$ to the number of turns $T$ in the response. By combining these two terms, we empirically incentivize the model to use more reflections during training for spatial navigation while not over-reflecting. We set $\alpha=0.2$ and $\gamma=0.05$ by default.

\section{Implementation Details} \label{sec:implementation_details}
\subsection{VRRL}
\textbf{Visual Grounding.} All training experiments are conducted on Qwen2.5-VL-3B-Instruct and Qwen2.5-VL-7B-Instruct using 4 NVIDIA A100 (80GB) GPUs. For the SFT stage, we train on 15K small table header lookup data with a learning rate of 5e-6 and a global batch size of 48. For the RL stage, We utilize a learning rate of 1e-6 and a global batch size of 32 on the SFT model that is trained for 5 epochs, where the model has already developed self-reflection output format. For GRPO, we sample $G=8$ rollouts per question and set the maximum trajectory length to $T=8$ turns. The KL coefficient is set to $\beta=0.01$. For the reward function, we set the shaping parameters $\sigma_1 = 2\delta_{\text{tol}}$ and $\sigma_2 = 5\delta_{\text{tol}}$ (where $\delta_{\text{tol}}=40$) to provide gradient signals at both coarse and fine granularities. We set the per-step cost to 0.
For the training curriculum, we set the mixing probability $\rho = 2/3$, meaning 66\% of training samples come from standard on-policy generation (with RTM applied) and 33\% from the replay buffer (Buffered Roll-In). 

The replay buffer $\mathcal{B}$ has a maximum capacity of 500 prefixes. To prevent the model from overfitting to failure states, which could cause the policy to unlearn immediate termination, we enforce a constraint where 30\% of the examples in $\mathcal{B}$ are forced to be \emph{correct} states (prefixes ending in a successful point). This ensures the policy practices validation (terminating when correct) alongside correction (refining when wrong). The buffer operates as a First-In-First-Out (FIFO) queue; given the small capacity, this ensures the ``mistake states'' remain relevant to the current policy's capabilities.

We run RL training for 1200 steps for all of our trained models and most models converge in 600 steps. We then use the checkpoints that achieve the best performance on a 200-example holdout set on the \emph{Large Table} task for all OOD evaluations.

\textbf{Spatial Navigation.} All training experiments are conducted on Qwen2.5-VL-3B-Instruct and Qwen3-VL-4B-Instruct using 4 NVIDIA A100 (80GB) GPUs. Following the settings and training configurations in \citet{xu2026visual}, we first warm-up the model with 10 epochs of SFT on the 3K single-turn examples of maps of size $3-5$ with only the direct answer as the task is very out-of-distribution of the instruction-tuned model. 

For the SFT stage of $\text{Single-SFT}$ and $\text{Multi-SFT}$, we train on 4K examples from maps of size $4-5$ as performance on $3\times3$ maps is already saturated after the warm-up stage. We run SFT with a learning rate of 2e-6 and a global batch size of 48. 

For the RL stage, we use 2K data from maps of size $4-5$. we apply \emph{online data filtering} adapted from DAPO \citep{yu2026dapo} to stabilize training by filtering out examples where all rollouts get 1 or 0 reward uniformly. For Qwen3-VL-4B, because its base capability is strong enough for multi-turn reflection, we use a smaller reflection bonus coefficient in the reflection reward to prevent reward hacking with $\alpha=0.1$ and $\gamma=0.01$.  We utilize a learning rate of 1e-6 for Qwen2.5-VL-3B and 5e-7 for Qwen3-VL-4B. We use a global batch size of 32 on the SFT model that is trained for 5 epochs. For GRPO, we sample $G=8$ rollouts per question and set the maximum trajectory length to $T=8$ turns. The KL coefficient is set to $\beta=0.01$. We use the same training curriculum and replay buffer configurations as the one for the visual grounding task. We use $\lambda_{\text{deg}}=0.5$ to balance between incentivizing reflections and penalizing over revisions.

We run RL training for 1200 steps for all of our trained models and most models converge in 250 steps given the strong warm-up-ed model. We then use the checkpoints that achieve the best performance on a 250-example heldout set on the in-distribution maps.

\subsection{Baselines}

\textbf{VL-Rethinker.} We directly evaluate the official released models of VL-Rethinker \citep{wang2026vlrethinker} with two model sizes (7B and 32B) on our evaluation tasks to test their OOD generalization capability when using textual CoT for self-reflection, since these models have already been trained on chart and diagram data during fine-tuning. At inference time, we follow \citet{wang2026vlrethinker} and prepend reflection prefixes after each assistant turn to encourage the model to reflect. We set the maximum number of reflections to $5$.

\textbf{Reflection Tuning.} \citet{wu2025guireflection} proposes reflection tuning to train models to self-reflect using visual feedback through online iterative SFT, where a teacher model provides supervision signals for revising incorrect generations. The original setup in \citet{wu2025guireflection} focuses on GUI tasks. It first performs SFT on multi-turn self-reflection data to obtain a warm-up model, and then applies iterative SFT on top of this model. To adapt this method to our setting, we use $\text{Multi-SFT}$ as the base model and apply 5 iterations of iterative SFT, following the same SFT-then-RL pipeline as our method. At each iteration, SFT data is created by sampling trajectories from the training prompts and collecting both correct and incorrect trajectories. For incorrect trajectories, we inject reflection by generating thinking traces together with the ground-truth answer, following \citet{wu2025guireflection}.

\section{Data Generation for Visual Grounding Tasks} \label{sec: sync_data_gen}

In this section, we describe the pipeline for synthesizing our training and evaluation datasets. The core advantage of programmatic generation is the ability to extract perfect ground-truth spatial metadata, which we leverage to construct precise visual question-answering pairs. Examples of each setting can be found in Figure~\ref{fig:ood_tasks}.

\subsection{Dataset Splits and Evaluation Tasks}

We train on a single in-distribution task, \emph{Table Lookup}, and evaluate on both held-out in-distribution examples and four OOD tasks that test different axes of generalization: scale, query type, and visual domain. Each test set contains 1K examples.

\textbf{Table Lookup: Training and In-distribution Evaluation.}
We generate small arXiv-style tables with 5 to 15 rows and columns across 15 academic domains. Queries ask the model to localize row or column headers, e.g., ``Find the column header `F1-Score'.'' This task serves as both the training distribution and the in-distribution evaluation setting.

\textbf{OOD-Large Tables: Size Generalization.}
We evaluate on substantially larger tables with 20 to 50 rows and columns, while preserving the same header-localization query format as the training task.

\textbf{OOD-Cell Query: Query-Type Generalization.}
We evaluate on small tables but query \emph{cell contents} (body text) rather than headers. This tests the model's ability to generalize the pointing mechanism to unseen query types without explicit training. Note that this represents a harder task: localizing a row or column header is effectively a 1D search problem along a single axis, whereas finding an exact inner cell requires precise 2D spatial localization.

\textbf{OOD-Bar Chart: Domain Generalization.}
We synthetically generate bar chart images with fine-grained control over chart metadata. This evaluates the model's ability to transfer its self-reflection skills to a novel visual domain where the objective is to point to the highest, lowest bar, or any bar given a specific category label. While visually identifying these bars is straightforward, we show that precisely localizing them within the image coordinate space remains highly error-prone (Table~\ref{tab:main_results}).

\textbf{OOD-Scatter Plot: Domain Generalization.}
We synthetically generate scatter plots where each plot contains 10 to 20 dots labeled with unique letters, and the model must point to the dot corresponding to a queried label. This reflects realistic data-extraction scenarios, where accurate spatial grounding of densely packed, uniquely labeled data points is an essential prerequisite for downstream tasks, such as reasoning over complex charts \citep{tang2025chartmuseum, masry-etal-2025-chartqapro}.

\subsection{Table Image Generation}

To generate table content, we sample terminology from 15 predefined academic domains, including machine learning, computer vision, natural language processing, reinforcement learning, and bioinformatics. For each domain, we define tailored vocabularies for column groups (e.g., dataset names like ImageNet or COCO), row groups (e.g., model architectures or size categories), evaluation metrics, and baseline methods. The cell values are populated using custom generators that simulate realistic numeric formats, such as percentages, floating-point numbers, and integer scores. We ensure that every data cell within a single table contains a unique value. This prevents spatial ambiguity during evaluation, ensuring there is only one correct answer per question.

We model four distinct table structures to reflect the layout diversity found in scientific literature:
\begin{enumerate}[leftmargin=10pt]
    \item \textbf{Single-level:} A basic table with a single row of column headers and a single column of row headers.
    \item \textbf{Multi-column:} Introduces hierarchical column groups, where an overarching category (e.g., a dataset) spans multiple sub-metrics.
    \item \textbf{Multi-row:} Introduces hierarchical groupings along the vertical axis, categorizing specific methods under broader taxonomies.
    \item \textbf{Fully hierarchical:} The most complex structure, combining multi-level spanning across both rows and columns.
\end{enumerate}

The visual rendering of the tables is implemented programmatically using Python. The output image dimensions dynamically scale to accommodate the generated table layout. During rendering, the absolute pixel coordinates of every individual cell, header, and spanning group are recorded. Leveraging this precise spatial metadata, we synthesize templated natural language question-answer pairs that require the model to locate specific elements, such as pointing to a specific row header, a nested column header, or an individual data cell.

\subsection{Bar Chart Image Generation}
We synthesize bar chart instances by independently sampling layout parameters for each example. 

\begin{enumerate}[leftmargin=10pt]
    \item \textbf{Overall layout:} The number of bars is drawn uniformly randomly between 5 and 10 bars. The height of the bar chart is randomly drawn between 512 and 768 pixels, and the width is fixed at a 16:9 aspect ratio relative to the sampled height. 
    \item \textbf{Category labels:} X-axis category labels are randomly sampled from several semantic schemas, including calendar months, weekdays, fiscal quarters, years, hours, business functions, fruits, and geographic regions, to ensure labels remain concise and contextually coherent. 
    \item \textbf{Bars:} Bar values are sampled i.i.d. from $\mathrm{Uniform}(0, 100)$. Bar colors are assigned by cycling through a fixed palette of 15 visually distinct hues, with the number of unique colors clamped to the sampled color count.
\end{enumerate}

Based on the layout above, SVG charts are rendered programmatically via a layout engine. Each SVG is subsequently rasterized to a PNG at the target resolution. Because all element positions are determined analytically, pixel-accurate ground-truth coordinates can be computed directly from the chart parameters without any post-hoc image processing. Question–answer pairs are synthesized from three types of templated natural language queries, including locating the bar with the largest or smallest value and identifying the bar corresponding to a named category. The models are prompted to answer the question by pointing to the exact pixel coordinate within the bounding box that defines the target bar. 

\subsection{Scatter Plot Image Generation}

Scatter plot generation begins by establishing a continuous two-dimensional coordinate system. For each plot, the visible data ranges for the x- and y-axes are determined by randomly cropping sub-spans from a broader, predefined global value range. A variable number of data points are then uniformly sampled as integer coordinates within these visible boundaries. 

To prevent visual occlusion and spatial ambiguity, we employ a rejection-sampling collision avoidance mechanism. This ensures that every plotted element, both the data points and their corresponding alphanumeric text labels, occupies a strictly unique coordinate pair. If overlaps are detected between points and labels, the coordinates are iteratively resampled.

\section{Data for Spatial Navigation Tasks} \label{sec:frozenlake_data}

We directly adopt the training and evaluation setup of FrozenLake from \citet{xu2026visual}, training on maps with sizes ranging from $3$–$5$ grids and evaluating on larger maps with sizes of $6$–$7$ grids to evaluate generalization. Each map size contains 250 examples in the evaluation set.

To induce self-reflection behavior through $\text{Multi-SFT}$ prior to RL training, we do not employ templated reasoning traces as used for visual grounding tasks, since \citet{xu2026visual} show that such templates are ineffective for training VLMs on this task. Instead, we distill CoT trajectories from GPT-5.4 \citep{singh2026openaigpt5card} that demonstrate corrections of randomly injected errors on the training maps and use them as self-reflective SFT data. Following \citet{xu2026visual}, we retain the same prompt format with `think', `answer', and `final' tags: when the model generates an answer proposal in the first turn, it emits an `answer' without justification (thinking). In the subsequent turn, the model needs to either finalize the answer with `final' tag (no justification needed), or revise with `think' and then a new answer proposal with `answer.' This scheme preserves the direct answer capabilities of the warm-up model in the first turn while still allowing models to perform reflection and revision. We further encourage reflection by prepending an instruction that asks the model either to revise its previous answer or terminate the trajectory after receiving visual feedback.

\section{Licenses} \label{sec:licenses}
We use the following publicly available datasets from prior works with open licenses.
\paragraph{FrozenLake} We access the FrozenLake data via Gym \citep{brockman2016openaigym} that uses the MIT license and data is available at: 
\url{gymnasium.farama.org}.

\section{Output Examples} \label{sec:example_outputs}

We show example outputs from our model on the OOD tasks.

\newpage
\begin{figure*}
\begin{imagebox}[]{Example: \emph{Large Table} Question (23 rows, 20 columns)}%
\end{imagebox}
\begin{modelsection}[]{Turn 1}
{\centering
\includegraphics[width=0.8\linewidth]{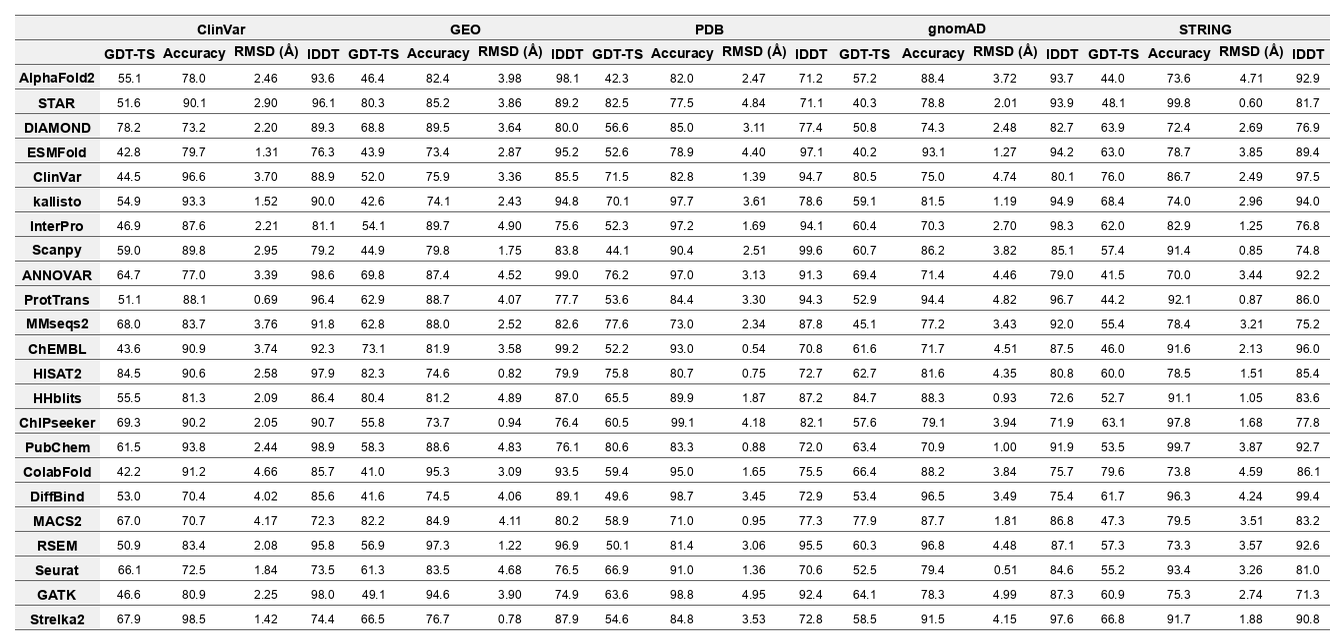}\par
}
[User] Question: Locate ``Accuracy'' in the ``PDB'' column group. Return its pixel coordinate in a list format [x\_coordinate, y\_coordinate].\\

[Assistant]
\begin{alltt}\small
```json
\{
  "think": "I need to locate `Accuracy' in the `PDB' column group. I'll point to its \\location.",
  "function_call": \{
    "name": "pointing",
    "arguments": \{
      "points": [
        \{
          "description": "Accuracy in the PDB column group",
          "x": 814,
          "y": 54
        \}
      ]
    \}
  \}
\}
```
\end{alltt}
\end{modelsection}
\caption{Example output for \emph{Large Table} question.} \label{fig:large_table_exp}
\label{fig:example-visual-1}
\end{figure*}
\clearpage
\begin{figure*}
\begin{modelsection}[]{Turn 2}
\par\noindent [User]\par
{\centering
\includegraphics[width=0.2\linewidth]{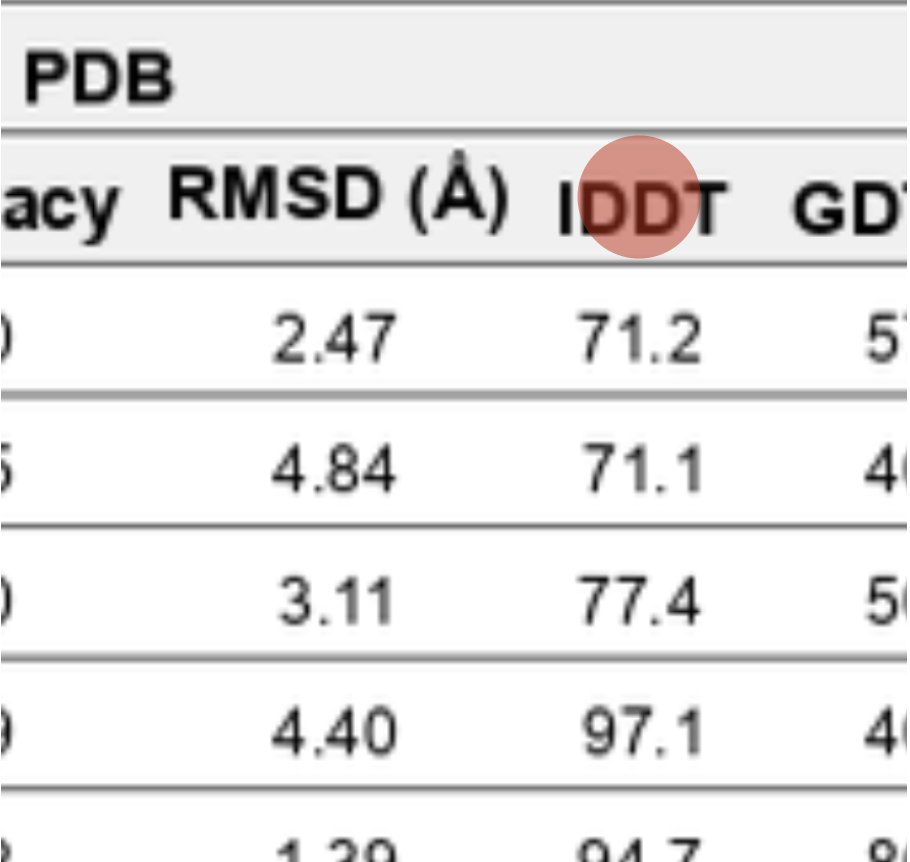}\par
}

[Assistant]
\begin{alltt}\small
```json
\{
  "think": "The pointer is not on `Accuracy' but on `IDDT'. I need to move to the left.",
  "function_call": \{
    "name": "pointing",
    "arguments": \{
      "points": [
        \{
          "description": "Accuracy in the PDB column group",
          "x": 683,
          "y": 54
        \}
      ]
    \}
  \}
\}
```
\end{alltt}
\end{modelsection}
\begin{modelsection}[]{Turn 3}
\par\noindent [User]\par
{\centering
\includegraphics[width=0.2\linewidth]{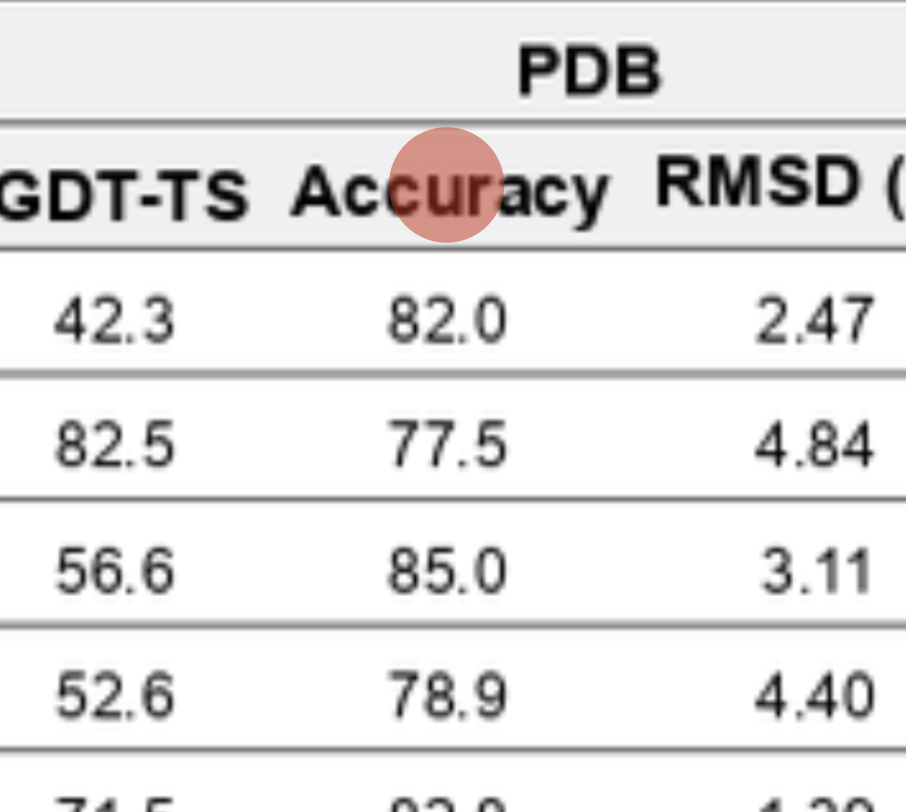}\par
}

[Assistant]
\begin{alltt}\small
```json
\{
  "think": "The image confirms that I'm pointing to `Accuracy`. I'll terminate now.",
  "function_call": \{
    "name": "terminate",
    "arguments": \{
      "answer": [683, 54]
    \}
  \}
\}
```
\end{alltt}
\end{modelsection}
\caption{Figure~\ref{fig:large_table_exp} continuation.}
\end{figure*}

\newpage
\clearpage
\begin{figure*}
\begin{imagebox}[]{Example: \emph{Bar Chart} Question (7 Bars)}%
\end{imagebox}
\begin{modelsection}[]{Turn 1}
{\centering
\includegraphics[width=0.8\linewidth]{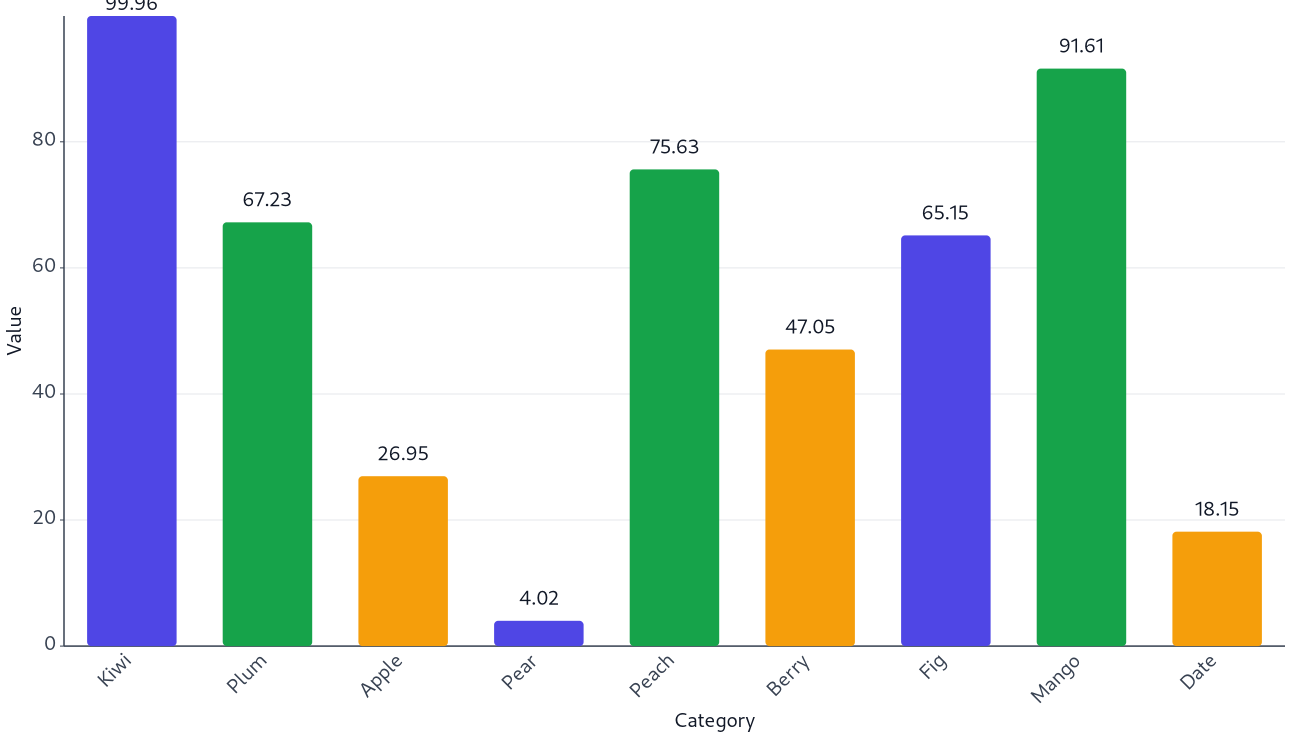}\par
}
[User] Question: Locate the bar where the category is ``Mango''. Return its pixel coordinate in a list format [x\_coordinate, y\_coordinate].\\

[Assistant]
\begin{alltt}\small
```json
\{
  "think": "I need to locate the bar for the category 'Mango'. I'll point to its \\location.",
  "function_call": \{
    "name": "pointing",
    "arguments": \{
      "points": [
        \{
          "description": "Mango",
          "x": 840,
          "y": 297
        \}
      ]
    \}
  \}
\}
```
\end{alltt}
\end{modelsection}
\caption{Example output for \emph{Bar Chart} question.} \label{fig:bar_chart_exp}
\label{fig:example-visual-bar-1}
\end{figure*}
\clearpage
\begin{figure*}
\begin{modelsection}[]{Turn 2}
\par\noindent [User]\par
{\centering
\includegraphics[width=0.2\linewidth]{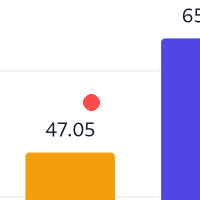}\par
}

[Assistant]
\begin{alltt}\small
```json
\{
  "think": "I'm pointing at 'Berry' instead of 'Mango'. Let me move right.",
  "function_call": \{
    "name": "pointing",
    "arguments": \{
      "points": [
        \{
          "description": "Mango",
          "x": 940,
          "y": 315
        \}
      ]
    \}
  \}
\}
```
\end{alltt}
\end{modelsection}

\begin{modelsection}[]{Turn 3}
\par\noindent [User]\par
{\centering
\includegraphics[width=0.2\linewidth]{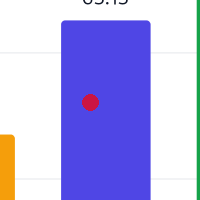}\par
}

[Assistant]
\begin{alltt}\small
```json
\{
  "think": "I'm pointing at 'Fig' instead of 'Mango'. Let me move right.",
  "function_call": \{
    "name": "pointing",
    "arguments": \{
      "points": [
        \{
          "description": "Mango",
          "x": 1007,
          "y": 334
        \}
      ]
    \}
  \}
\}
```
\end{alltt}
\end{modelsection}
\caption{Figure~\ref{fig:bar_chart_exp} continuation.}
\label{fig:bar_chart_exp_cont}
\end{figure*}

\newpage 
\begin{figure*}

\begin{modelsection}[]{Turn 4}
\par\noindent [User]\par
{\centering
\includegraphics[width=0.2\linewidth]{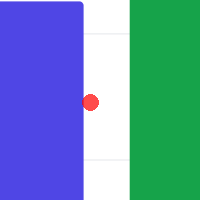}\par
}

[Assistant]
\begin{alltt}\small
```json
\{
  "think": "I'm pointing at 'Fig' instead of 'Mango'. Let me move right.",
  "function_call": \{
    "name": "pointing",
    "arguments": \{
      "points": [
        \{
          "description": "Mango",
          "x": 1080,
          "y": 352
        \}
      ]
    \}
  \}
\}
```
\end{alltt}
\end{modelsection}

\begin{modelsection}[]{Turn 5}
\par\noindent [User]\par
{\centering
\includegraphics[width=0.2\linewidth]{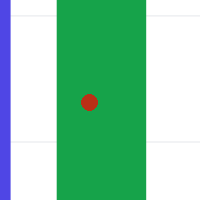}\par
}

[Assistant]
\begin{alltt}\small
```json
\{
  "think": "The crop now correctly shows the bar for the category 'Mango'.",
  "function_call": \{
    "name": "terminate",
    "arguments": \{
      "answer": [1080, 352]
    \}
  \}
\}
```
\end{alltt}
\end{modelsection}
\caption{Figure~\ref{fig:bar_chart_exp_cont} continuation.}
\end{figure*}

\newpage
\clearpage
\begin{figure*}
\begin{imagebox}[]{Example: \emph{FrozenLake} Question ($6\times6$ map)}%
\end{imagebox}
\begin{modelsection}[]{Turn 1}
{\centering
\includegraphics[width=0.5\linewidth]{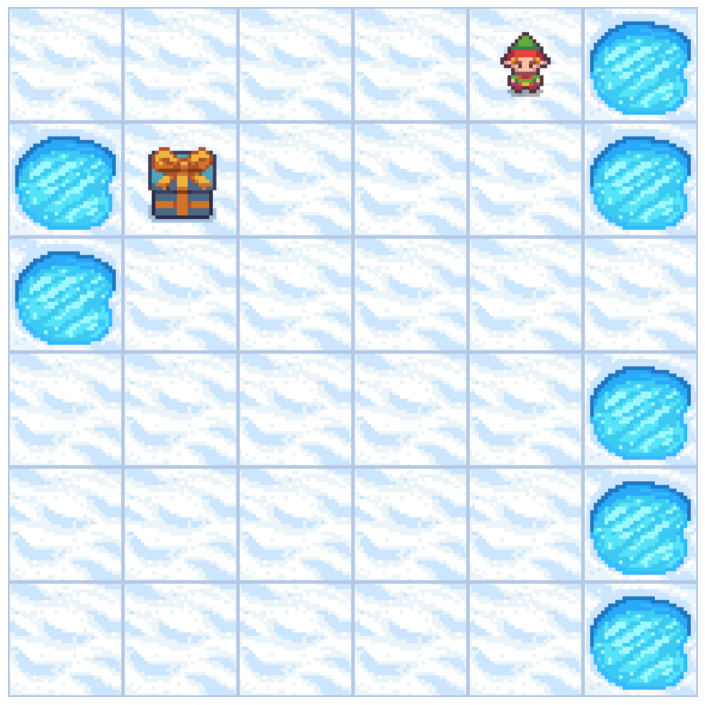}\par
}
[User] Question: Generate the shortest action sequence.\\

[Assistant]
\begin{alltt}\small
<answer> left left down </answer>
\end{alltt}
\end{modelsection}
\caption{Example output for \emph{FrozenLake} question.} \label{fig:frozen_lake_exp}
\label{fig:example-visual-frozenlake-1}
\end{figure*}
\clearpage
\begin{figure*}
\begin{modelsection}[]{Turn 2}
\par\noindent [User]\par
{\centering
\includegraphics[width=0.2\linewidth]{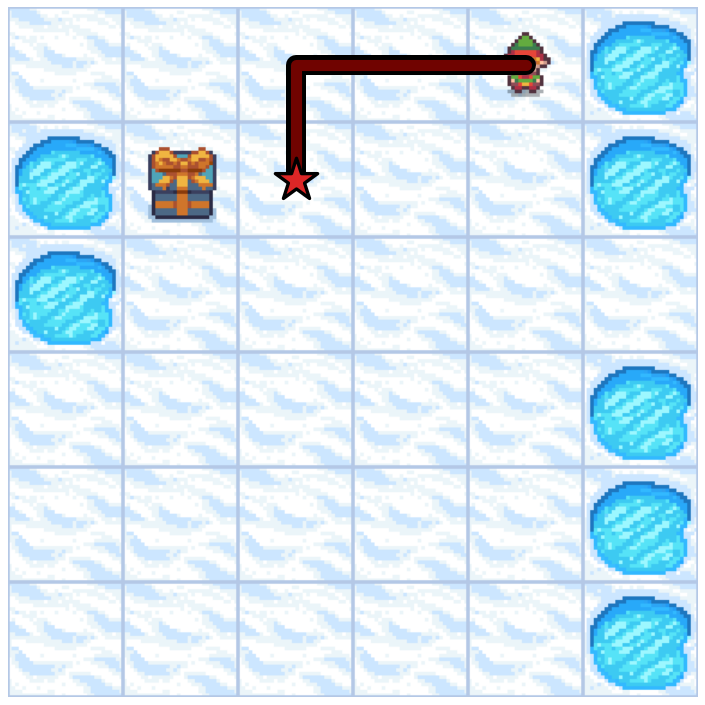}\par
}

[Assistant]
\begin{alltt}\small
<think> In the feedback image, the red path ends on a snow cell directly beneath the elf, 
with the red star sitting on that spot. The first two moves were correct, but the third 
move is the first incorrect step because it goes off the grid. The gift is still in the 
upper-left corner, and the elf is already on a safe snow cell. To fix it, continue along 
the same line toward the gift, then move one cell to the side to reach the gift. </think>
<answer>left left left down</answer>
\end{alltt}
\end{modelsection}

\begin{modelsection}[]{Turn 3}
\par\noindent [User]\par
{\centering
\includegraphics[width=0.2\linewidth]{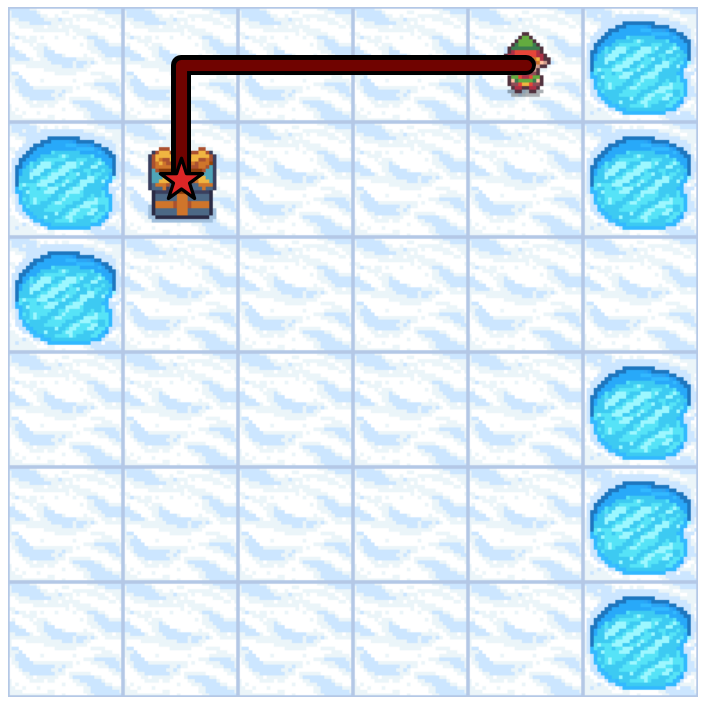}\par
}

[Assistant]
\begin{alltt}\small
<Final> left left left down </Final>
\end{alltt}
\end{modelsection}
\caption{Figure~\ref{fig:frozen_lake_exp} continuation.}
\label{fig:frozen_lake_exp_cont}
\end{figure*}

\end{document}